\documentclass[journal]{IEEEtran} 
\IEEEoverridecommandlockouts
\usepackage{cite}
\usepackage{amsmath,amssymb,amsfonts}
\usepackage{algorithmic}
\usepackage{graphicx}
\usepackage{textcomp}
\usepackage{float}
\usepackage{xcolor}
\usepackage{subfigure}

\def\BibTeX{{\rm B\kern-.05em{\sc i\kern-.025em b}\kern-.08em
    T\kern-.1667em\lower.7ex\hbox{E}\kern-.125emX}}
\usepackage{booktabs}
\begin{document}

\title{Parkinson’s Disease Detection via Resting-State Electroencephalography Using Signal Processing and Machine Learning Techniques\\

}

\author{
    \IEEEauthorblockN{Krish Desai}
    \\
    \IEEEauthorblockA{\textit{Leland High School, San Jose, United States}}\\
    \IEEEauthorblockA{Corresponding author: Krish Desai (krishdesaiedu@gmail.com)}
    \IEEEauthorblockA{ORCID ID: 0000-0003-2559-8910}
}

\maketitle

\begin{abstract}
    Parkinson’s Disease (PD) is a neurodegenerative disorder resulting in motor deficits due to advancing degeneration of dopaminergic neurons. PD patients report experiencing tremor, rigidity, visual impairment, bradykinesia, and several cognitive deficits. Although Electroencephalography (EEG) indicates abnormalities in PD patients, one major challenge is the lack of a consistent, accurate, and systemic biomarker for PD in order to closely monitor the disease with therapeutic treatments and medication. In this study, we collected Electroencephalographic data from 15 PD patients and 16 Healthy Controls (HC). We first preprocessed every EEG signal using several techniques and extracted relevant features using many feature extraction algorithms. Afterwards, we applied several machine learning algorithms to classify PD versus HC. We found the most significant metrics to be achieved by the Random Forest ensemble learning approach, with an accuracy, precision, recall, F1 score, and AUC of 97.5{\%}, 100{\%}, 95{\%}, 0.967, and 0.975, respectively. The results of this study show promise for exposing PD abnormalities using EEG during clinical diagnosis, and automating this process using signal processing techniques and ML algorithms to evaluate the difference between healthy individuals and PD patients. 
\end{abstract}


\section{Introduction}
Parkinson’s disease (PD) is a fast-growing neurodegenerative disease resulting from several pathophysiological conditions, such as a-synuclein aggregation, disruption of lysosomes, neuroinflammation, disruption of synaptic transport issues, dysfunction of mitochondria, and neuronal death [1]. Cellular processes in patients with PD are typically influenced by environmental and genetic factors [2]. PD patients experience many symptoms, such as tremor, rigidity, akinesia, bradykinesia, postural instability, and several cognitive deficits [3]. Other associated features for PD patients are sleep dysfunction, a loss of smell, mood disorders, periodic limb movements during REM sleep, and excess salivation [4]. Identifying PD at an early stage is difficult, as an accurate and timely diagnosis presents several challenges [5]. Recent studies suggest that prodromal PD disease is currently diagnosed with routine work-up clinical protocols, analysis of genetic subtypes, genetic and imaging tests, and even fluid markers [5]. However, current clinical diagnosis is unsystematic and doesn’t provide promise for early detection of PD. Additionally, incorrect clinical diagnosis is common since several conditions mimic PD, such as dementia with Lewy bodies, Multiple System Atrophy, Progressive Supranuclear Palsy, Corticobasal Syndrome, Essential Tremor, and Normal Pressure Hydrocephalus [6]. A systematic diagnostic plan using clinical factors would be pertinent for slowing down the progression of PD, and although there is no cure, several treatments exist. Surgical procedures such as deep brain stimulation (DBS) or MRI guided focused ultrasound and medications — Carbidopa-levodopa, dopamine agonists, MAO B inhibitors, and Catechol O-methyltransferase (COMT) inhibitors —can manage PD symptoms such as tremors, poor sleep, or other conditions [7]. \newline
\indent Assisting a systematic clinical diagnosis could be neuroelectric recordings, serving as an integral biomarker in many stages of PD. Han et al. (2013) found that changes in neural activity recorded in an Electroencephalogram (EEG), a non-invasive record of electrical activity in cortical brain regions, can serve as a biomarker for neurodegenerative diseases such as PD[8]. Machine learning, an automated form of clinical data analysis, shows promise for identifying non conventional features in the time and frequency series analysis of Electroencephalographic signals through signal processing [9]. 

\section{Background Literature}
\indent Currently, several biomarkers are used for early PD diagnosis, such as imaging, biochemical, clinical, and genetic [10]. Imaging biomarkers include magnetic resonance imaging (MRI), optical coherence tomography (OCT), and even molecular techniques such as fluorodopa positron emission tomography (F-DOPA PET) [11]. Biochemical biomarkers typically include cerebrospinal fluid (CSF) exploration in cortical regions, neuroinflammatory reactions due to dopaminergic neuron loss, and neurosin presence in northern blotting [11]. Clinical biomarkers include motor impairment symptoms, such as incontinence, walking difficulty, and other irregularities while genetic biomarkers include mutations in DJ1, PINK1, Parkin, LRRK2, SNCA and GBA [11].\newline
\indent Several criterias, protocols, and tests are designed for diagnosing PD patients in the clinical scene. The Movement Disorder Society Clinical Diagnostic Criteria (MDS) focuses on non motor and motor manifestations in prodromal PD [12]. The Mini-Mental State Examination (MMSE) can also be used to diagnose PD by evaluating cognition, memory, language, executive function, and visuospatial skills [13]. \newline
\indent Also, examining the electrical activity in cortical regions is a technique used to identify PD. EEG, a measure of neuroelectric activity through electrodes placed around the scalp, is considered an important PD diagnostic tool used to measure cerebral information [14]. More importantly, EEG is an essential electrophysiological indicator of cognitive decline, a common effect of PD [15]. This quantitative approach to measuring cortical activity is useful for PD diagnosis, which can be streamlined with automated machine learning algorithms.

\section{Materials and Methods}
This section describes the method for processing EEG signals and PD/HC classification, including data formatting, signal preprocessing, feature extraction, and model construction, performed primarily using MNE (https://mne.tools/) and SciPy (https://scipy.org/). A high-level overview of the several stages of signal processing and machine learning modeling for healthy individuals and Parkinson’s patients is shown in Figure 1. The raw EEG signals are initially read and the EEG signals from all 40 electrodes are averaged. The EEG signals are then resampled to a common frequency and cleaned using a band-pass filter to locate the optimal frequency region. After cleaning, EEG signals are divided into equal segments before HC/PD features are extracted, occasionally through further EEG decomposition, using several metrics: power spectrum, energy, mean, standard deviation, spectral entropy, Hjorth’s mobility, Hjorth’s activity, approximate entropy. Finally, different classifiers, such as Random Forest (RF), Support Vector Machine (SVM), Extra Trees (ET), and K-Nearest Neighbor (KNN), are employed to distinguish HC characteristics from those exhibited by PD individuals. 

\subsection{Dataset Overview}
Data collected for this study came from UC San Diego’s Resting State EEG Data for healthy control and Parkinson’s patients [16][17][18][19]. Data comprises resting state EEG recordings from 15 PD patients (mean age= 63.2±8.2 years) on and off dopaminergic medications and 16 Healthy Control (HC) patients (mean age= 63.5±9.6 years). Table 1 includes demographics for Parkinson’s patients and HC individuals belonging to this dataset. As shown by the Mini-Mental State Exam (MMSE) and the North American Adult Reading Test (NAART) that all participants took, the age range, gender, and handedness seemed to be similar across both the HC and PD groups. A movement disorder specialist from the Scripps Clinic in La Jolla, California diagnosed all patient’s (n=31) in this study. Participants provided consent in accordance to the Declaration of Helsinki and the Institutional Review Board of the University of California, San Diego. On-medication patients continued their medication course up until the session while off-medication patients discontinued medication use at least 12 hr before a session. EEG data were obtained using a 32-channel BioSemi ActiveTwo System, with a sampling frequency of 512 Hz. Eye blinks and movements were monitored using electrodes placed lateral to the patient’s left eye and participants focused their attention on a cross during the session. The resting state EEG data was recorded for 3 minutes. Each EEG signal was recorded from 32 EEG electrodes: Fp1, AF3, F7, F3, FC1, FC5, T7, C3, CP1, CP5, P7, P3, Pz, PO3, O1, Oz, O2, PO4, P4, P8, CP6, CP2, C4, T8, FC6, FC2, F4, F8, AF4, Fp2, Fz, and Cz. Two electrodes, placed over the right and left mastoids, were used as reference electrodes. Fp1, AF3, F7, F3, FC1, FC5, FC2, F4, F8, AF4, Fp2, Fz electrodes were placed in the frontal lobe region; P7, P3, Pz, PO3, O1, Oz, O2, PO4, P4, P8 electrodes were placed in the parietal lobe region; T7, T8 electrodes were placed in the temporal lobe region; C3, CP1, CP5, CP6, CP2, C4, Cz electrodes were placed in the central region of the brain.

\begin{table}[ht]
    \caption{Control and patient participant demographics} 
    \renewcommand{\arraystretch}{1.8} 
    \centering 
    \begin{tabular}{c c c} 
        \hline\hline 
        \textbf{Category} & \textbf{HC} & \textbf{PD} \\ [0.5ex] 
        \hline 
        Number of Participants & 16 & 15 \\ 
        Age (years) & $63.5\pm9.6$ & $63.2\pm8.2$ \\ 
        Sex & 9F/7M & 8F/7M \\ 
        Handedness & All R & All R \\ 
        NAART & $49.12\pm7.1$ & $46\pm6.27$ \\
        MMSE & $29.1\pm1.1$ & $28.9\pm1$ \\
        UPDRS On medication & N/A & $33.68\pm10.86$ \\ 
        UPDRS Off medication & N/A & $41.5\pm12.95$ \\ [1ex] 
        \hline 
    \end{tabular}
    \label{tab:participant_char} 
\end{table}

\subsection{EEG Signal Processing}
First, EEG data per patient was separated from the rest of the dataset and configured in the BDF file format. Time-series voltage channels were averaged for all EEG signals recorded from 40 electrodes to improve the signal-to-noise ratio and reduce susceptibility to artifacts. After the EEG data was converted into an MNE object, all EEG signals were resampled to a lower common frequency, 250 Hz, to improve signal quality.\newline
\indent A non-casual bandpass filter was applied to find the frequency region of interest, since PD neural activity was previously correlated to the 1-30 Hz frequency region [20]. Each patient’s EEG signal was split into five frequency bands: delta ($\Delta$), theta ($\Theta$), alpha ($\alpha$), beta ($\beta$), gamma ($\gamma$). The filtered frequency ranges for delta, theta, alpha, beta, and gamma were 0.5-4.5 Hz, 4.5-8.5 Hz, 8.5-11.5 Hz, 15.5-30 Hz, and 30-45 Hz, respectively. Firwin’s non-casual bandpass method was applied to filter out high-frequency regions and exclude slow drifts in the signal [21]. The characteristics of this one-pass, zero-phase bandpass were a lower and upper transition bandwidth of 0.50 Hz (-6 dB cutoff frequency), a hamming window with 0.0194 passband ripple, and a 53 dB stopband attenuation. The filtered EEG signals were then segmented into 5s epochs with a 1s overlap, and for each patient, the bad epochs were dropped. An example of a filtered EEG signal for each band is shown for an HC individual and PD patient

\begin{figure}[H]
    \centering
    \includegraphics[width=\textwidth]{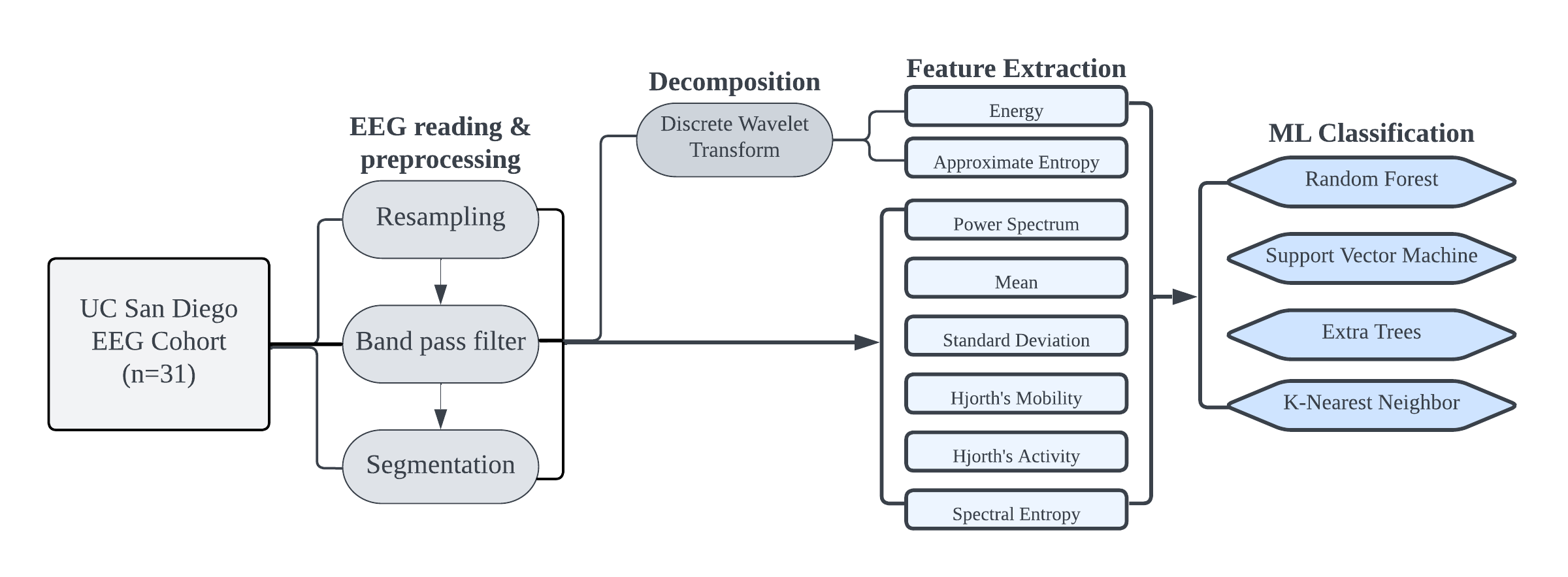}
    \caption{Diagram of proposed PD diagnostic method using EEG Signals}
    \label{fig:my_label}
\end{figure}

in Figure 2. As can be seen, HC individuals generally have lower microvolt amplitudes compared to PD patients.

\begin{figure}[H]
    \centering
    \subfigure(A){\includegraphics[width=0.45\textwidth]{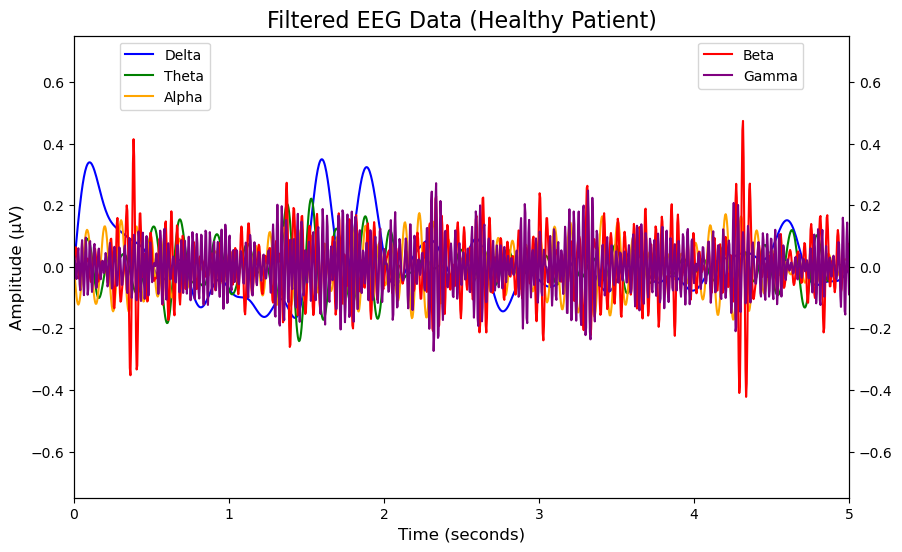}}
    \hfill
    \subfigure(B){\includegraphics[width=0.45\textwidth]{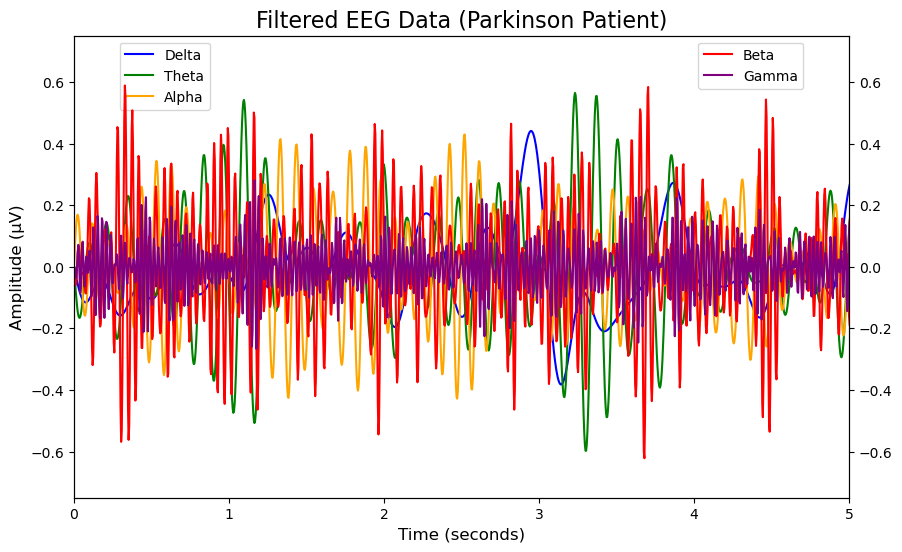}}
    \caption{Comparison of filtered EEG signals between a healthy patient and a Parkinson's patient. (A) Healthy Patient; (B) Parkinson's Patient}
    \label{fig:my_label}
\end{figure}

\indent After data formatting and EEG signal preprocessing, the features could be extracted. EEG signals are complicated and store a large amount of data. Therefore, the ability to extract relevant and proper features from any EEG signal is useful for any successful ML algorithm. The feature extraction stage in this study aims to create a low-dimensional, compact space using preprocessed EEG signals, and past literature suggests that many feature extraction methods have been proposed on
\\
\\
\\
\\
\\
\\
\\
\\
\\
\\
\\
\\
\\
\\
\\
\\
\\
\\
\\
the basis of particular objectives, including time domain series, frequency domain series, and nonlinear [22]. The following study extracts features on a per-epoch basis, including time domain features — mean and standard deviation—, frequency domain features — power spectrum and spectral entropy —, time-frequency domain features — energy —, and nonlinear features — Hjorth’s mobility, Hjorth’s activity, and approximate entropy —. Since the preprocessed EEG data is given in time series, features from this domain were extracted first, such as mean, standard deviation, and the energy of each signal. Mean and standard deviation aren’t necessarily complex, but they provide the dispersion and central tendency of the data which could be important [23]. Additionally, time domain EEG analysis is important when measuring synchronicity, a measure of how similar EEG signals are [24]. After time domain feature extraction, the frequency components of each EEG signal were analyzed by converting time series (amplitude vs time) to frequency series (amplitude vs frequency). Frequency analysis is used to measure the occurrence of events at a specific time, and it provides better spatial information for EEG analysis [24]. In this study, the Fast Fourier Transform (FFT) method was heavily used to analyze EEG signals in the frequency domain. FFT is applied to specific time intervals of EEG signal data, and each epoch is multiplied by a proper windowing function, allowing for periodograms to be calculated on a per-segment basis for each epoch [24]. Before analyzing frequency domain feature extraction methods, spectrograms, a frequency domain representation using FFT, were used to analyze the distributions in frequency across all five filtered bands. The spectrogram for this study is shown in Figure 3. \newline
\indent The frequency domain power-spectral density (PSD) leverages FFT and represents the distribution of power over its frequency components via sets of frequency bins, with each bin representing a range of frequencies [25][26]. Welch’s method of calculating PSD is useful when estimating the power spectrum of determined time sequences and increasing the resolution of power-spectral density [27]. Welch’s method creates periodograms on a per-segment basis by dividing the epoched signal into successive blocks and averaging to estimate the power spectra [28]. Welch’s power spectra weighs each segment with a window function, allowing for an estimate with less spectral leakage [28]. Equation 1 shows Welch’s method. 

\begin{align}
    P_{SD}^{Welch}(x(\omega_k)) \triangleq \frac{1}{K} \sum_{m=0}^{K-1} P_{x_m(M)}(\omega_k)
\end{align}

\indent Before the electrode channels were averaged, an example of PSD over frequency domain for HC/PD EEG signals is shown in Figure 4. In this particular study, FFT was applied to 1250 segments, with 250 segments of overlap, for each epoch and the hamming window parameter was passed. After PSD values were collected, spectral entropy, the measure of power-spectral density complexity, was calculated. First, PSD was normalized to ensure that the power spectrum was independent from the signal’s energy, and then spectral entropy was measured by multiplying the per-epoch natural logarithmic by the sum of the normalized power-spectral density values. 

\begin{figure}
    \centering
    \subfigure{\includegraphics[width=0.45\textwidth]{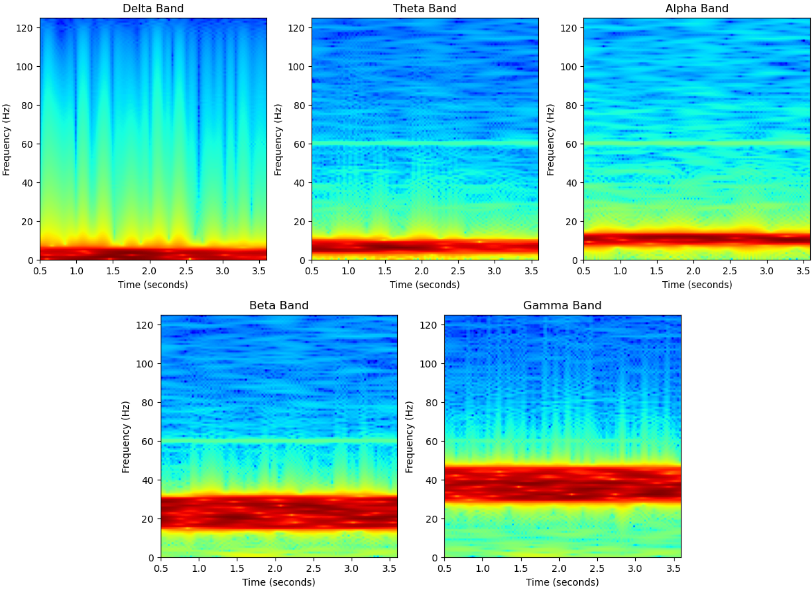}}
    \caption{Spectrogram of five frequency bands}
    \label{fig:my_label}
\end{figure}

\indent Along with time domain and frequency domain, time-frequency analysis of EEG can provide more information about neural synchrony and reveal how the phase angles of frequencies synchronize across space and time [29]. Further EEG decomposition was required to extract features in the time-frequency series. Daubechies Discrete Wavelet Transform (DWT) is capable of decomposing EEG signals into a set of mutually orthogonal components using a single function called the mother wavelet, allowing for analysis across both time and frequency domains [14]. After FFT transforms the signal to frequency series, Daubechies DWT uses low and high pass filters in the first level of decomposition to produce approximation (cA) and detail (cD) coefficients. The selected mother wavelet for decomposition is db4, and it is the most widely used mother wavelet for EEG signal analysis according to several review studies [30][31]. Daubechies DWT and coefficients are shown in Equation 2. A couple wavelet features were calculated using PyWavelets (https://pywavelets.readthedocs.io/), but only one was calculated in time-frequency series, which was energy. Energy of each epoch was calculated by taking the sum of the squares of the cA and cD coefficients.

\begin{figure}
    \centering
    \subfigure(A){\includegraphics[width=0.45\textwidth]{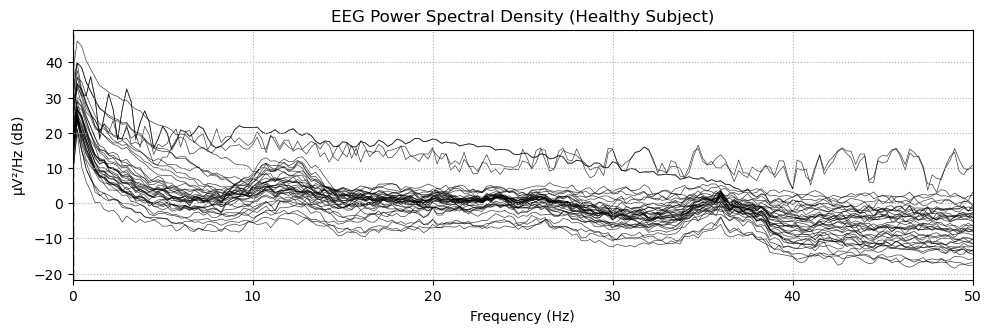}}
    \hfill
    \subfigure(B){\includegraphics[width=0.45\textwidth]{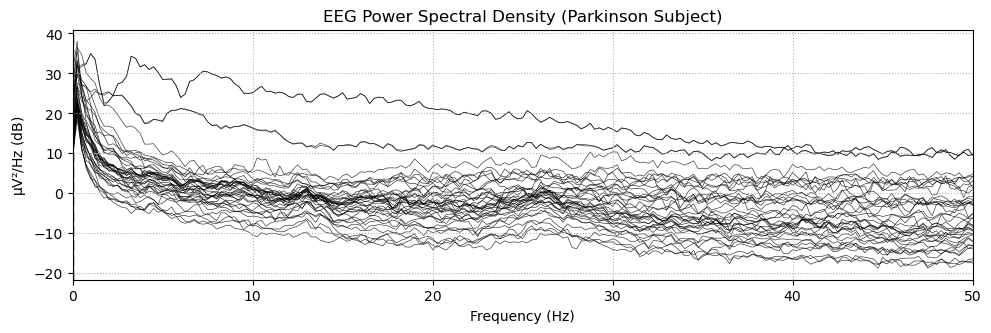}}
    \caption{Comparison of filtered EEG signals between a healthy patient and a Parkinson's patient. (A) Healthy Patient; (B) Parkinson's Patient}
    \label{fig:my_label}
\end{figure}

\begin{subequations}
\begin{center}
\begin{align}
    x(t) = \sum_{j} c_j(k) \varphi(t-k) + \sum_{k}\sum_{j-1} d_j(k) \varphi\left(2^{-j}t-k\right) \\
    c_j(n) = \sum_k x(n)\cdot h(2n-k) \\
    d_j(n) = \sum_k x(n)\cdot g(2n-k)
\end{align}
\end{center}
\end{subequations}

 Non-linear feature extraction methods can provide information regarding complex and dynamic interactions in brain regions and PD abnormalities which cannot be captured via EEG linear methods, such as non-stationary or nonlinear components [32]. Daubechies DWT can still be used for some non-linear feature extraction methods, since the wavelet transform can identify transient events such as sharp waves and spikes popular in PD abnormalities. Thus, the cA coefficients were used to calculate approximate entropy, a non-linear feature measuring complexity and randomness of cA, by taking the negative logarithm of the probability of finding similar patterns in the cA data.
Aside from wavelet transform, two more non-linear features were extracted to analyze the spatial and temporal properties of each signal. Hjorth’s parameters, activity and mobility, were calculated to describe the variance in each signal. Hjorth’s activity measures variance in time domain and surface of power spectrum over the frequency domain [33]. Hjorth’s mobility indicates the mean frequency in the time domain and represents the proportion of standard deviation over the power spectrum. Hjorth’s Activity and Mobility were calculated in Equation 3.

\begin{subequations}
\begin{center}
\begin{align}
    \text{Activity} = \operatorname{var}[y(t)] \\
    \text{Mobility} = \sqrt{\frac{\operatorname{var}\left(\frac{d y(t)}{dt}\right)}{\operatorname{var}[y(t)]}}
\end{align}
\end{center}
\end{subequations}

\subsection{Model Construction}

\indent In this study, classical machine learning algorithms were used to classify HC individuals and PD patients based on extracted EEG features. All machine learning algorithms were implemented using Scikit-learn (https://scikit-learn.org/), including Random Forest, Support Vector Machine, Extra Trees, and K-Nearest Neighbor. A random grid search was used to hyperparameter-tune all four classifiers. \newline
\indent Random Forest is an ensemble method that uses several randomized decision trees, each trained on different sub-samples of the data and different divisions of the features at each tree split point, and takes the average of their results. This approach allows the model to fix errors in preceding trees at node split points, allowing for improved accuracy, reduced overfitting, and less variability [34]. \newline
\indent Extra Trees (Extremely Randomized Trees) is another ensemble method which, in comparison to Random Forest, also recursively selects random subsets of data and features at each tree’s split point [35]. Unlike Random Forest, Extra Trees uses random split points, since at each candidate split point, the model selects a threshold value rather than choosing the best split point, allowing for a more diverse, random set of trees. Extra Trees random split point algorithm allows for reduced overfitting, similar to Random Forest’s greedy algorithm. \newline
\indent Support Vector Machines (SVM) uses hyperplanes for classification problems to partition data into two classes, 0 or 1. The SVM algorithm operates in an N-dimensional space, where N is the number of features. Each feature is represented as an axis in this space, and each data point corresponds to a point in this N-dimensional plane. The algorithm differentiates data points belonging to two different classes via a hyperplane and minimizes the functional margin, the distance between the hyperplane and the closest data points in each class [36]. \newline
\indent K-Nearest Neighbor is a supervised learning algorithm that calculates the Euclidean distance, a space between the test data and every training point [37]. K represents the points closest to the test data, and the Euclidean distance between every K and the test points. The probability of the test data belonging to the classes of K training data represents the KNN algorithm. KNN doesn’t require assumptions about data distribution, it’s non-parametric, and it only considers K nearest neighbors, rather than the whole dataset, making it a widely used model due to simplicity.

\subsection{Performance Evaluation}
\indent We use a k-fold cross validation (CV) technique to achieve reliable performance metrics. We used k=10 for all four classifiers, with 90{\%} of the feature data used for training and 10{\%} used for testing. Because k=10, 9 partitions of the features were used to train the model while 1 partition was used to test the model for performance. The cross validation procedure is carried out k number of times, and to measure the classifiers performance, the results of the ten cross-validation rounds are averaged. \newline
\indent Several metrics are used to evaluate the performance for the four classifiers: accuracy, precision, recall, F1 score, and receiver operating characteristic (ROC) curve. Accuracy is simply calculated using the following equation, where TN= {\#} of True Negatives, TP= {\#} of True Positives, FN={\#} of False Negatives, and FP={\#} of False Positives.\newline
\\
Accuracy Equation:
\begin{align}
    \text{Accuracy} = \frac{TN+TP}{TN+TP+FN+FP} \times 10
\end{align}
\\
Precision is calculated using TP and FP in the following formula:
\begin{align}
    \text{Precision} = \frac{TP}{TP+FP} \times 100\%
\end{align}
\\
Recall is calculated through the following formula:
\begin{align}
    \text{Recall} = \frac{TP}{TP+FN}
\end{align}
\\
F1-score is computed using the following equation:
\begin{align}
    \text{F1-score} = 2 \times \frac{\text{Precision} \times \left(\frac{TP}{TP+FN}\right)}{\text{Precision}+\left(\frac{TP}{TP+FN}\right)} \times 100\%
\end{align}
\\
\indent The ROC curve is a graphical representation showing the trade-off between the model’s specificity and sensitivity as the discrimination threshold is varied. The ROC curve plots the true positive rate (TPR) against the false positive rate (FPR), where TPR measures the accurate identification of positive instances, while FPR measures the misclassification of negative instances as positive. The area under the ROC curve (AUC) is a common metric used to measure the area of unit square, with values ranging from 0 to 1. The closer the AUC value is to 1, the more accurate and better the classifier is. 

\section{Results}
\subsection{Model Results}
\indent Cross validation was used to split the data into training and testing sets. We used a cross validation random grid search algorithm to find the optimal hyperparameters for each classifier shown in Table 2. 

\begin{table}[ht]
    \caption{Grid search tuned hyperparameters}
    \renewcommand{\arraystretch}{1.5}
    \centering
    \begin{tabular}{c p{5cm}}
        \hline\hline
        \textbf{Classifier} & \textbf{Hyperparameters} \\ [0.5ex]
        \hline
        Random Forest & No. of estimators=1000, max leaf nodes=100, max depth=10 \\
        \hline
        Support Vector Machine & Kernel=linear, \newline shrinking=False \\
        \hline
        Extra Trees & No. of estimators=1000, max leaf nodes=100, max depth=10 \\
        \hline
        K-Nearest Neighbor & No. of neighbors=1000 \\ [1ex] 
        \hline
    \end{tabular}
    \label{tab:participant_char} 
\end{table}

\indent The results for the grid-search tuned models were visualized in a variety of ways. For each classifier, the true-positives (TP), true-negatives (TN), false-positives (FP), and false-negatives (FN) were collected from the prediction arrays to calculate accuracy, precision, recall, F1 score, and area-under-the-curve (AUC). The calculated metrics are shown in Table 3. 

\begin{table}[ht]
    \caption{Classifier Metrics}
    \renewcommand{\arraystretch}{1.3}
    \centering
    \begin{tabular}{p{1cm}|p{1cm}|p{1cm}|p{1cm}|p{1cm}|p{1cm}}
        \hline\hline
        \textbf{Classifier} & \textbf{Accuracy ({\%})} & \textbf{Precision ({\%})} & \textbf{Recall ({\%})} & \textbf{F1 Score} & \textbf{ROC AUC} \\ [0.5ex]
        \hline
        Random Forest & 97.5 & 100 & 95 & 0.967 & 0.975 \\
        \hline
        SVM & 35 & 16.667 & 50 & 0.250 & 0.5 \\
        \hline
        Extra Trees & 94.167 & 96.667 & 95 & 0.947 & 0.925 \\
        \hline
        KNN & 29.167 & 5 & 5 & 0.05 & 0.3 \\ [1ex] 
        \hline
    \end{tabular}
    \label{tab:participant_char} 
\end{table}

AUC plots, created by plotting the true positive rate (TPR) against the false positive rate (FPR), were used to visualize the ROC curve. A precision-recall curve was also used to visualize the trade-off between precision and recall for different threshold values. Both the ROC curve and the precision-recall curve are shown in Figure 5. Since Random Forest and Extra Trees yielded the best metrics and showed the best performance based off both curves, two confusion matrices were created to show TP, TN, FP, and FN for these models in Figure 6.

\begin{figure}[H]
    \centering
    \subfigure(A){\includegraphics[width=0.45\textwidth]{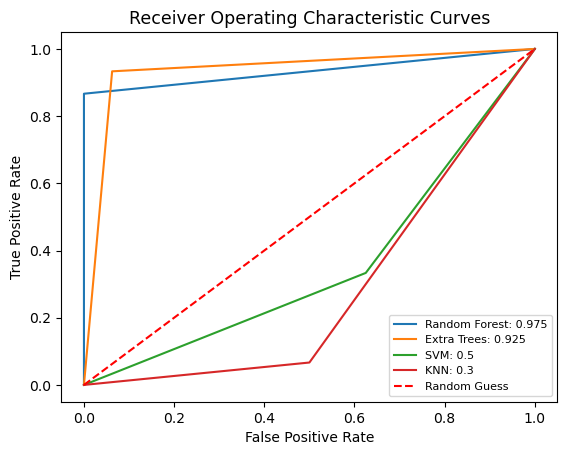}}
    \hfill
    \subfigure(B){\includegraphics[width=0.45\textwidth]{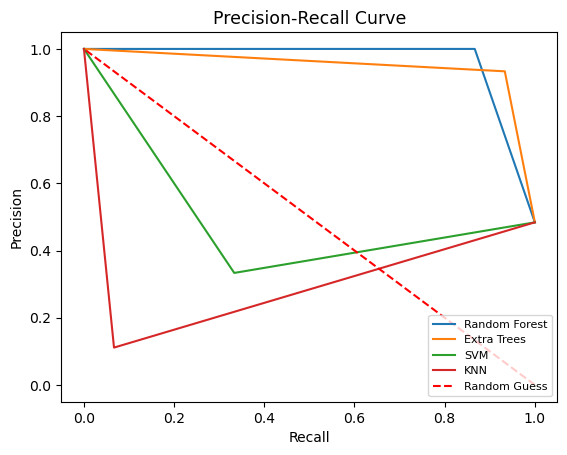}}
    \caption{Confusion Matrices showing predicted versus actual labels for Random Forest and Extra Trees. (A) ROC curve; (B) Precision-Recall curve}
    \label{fig:my_label}
\end{figure}

\begin{figure}[H]
    \centering
    \subfigure(A){\includegraphics[width=0.45\textwidth]{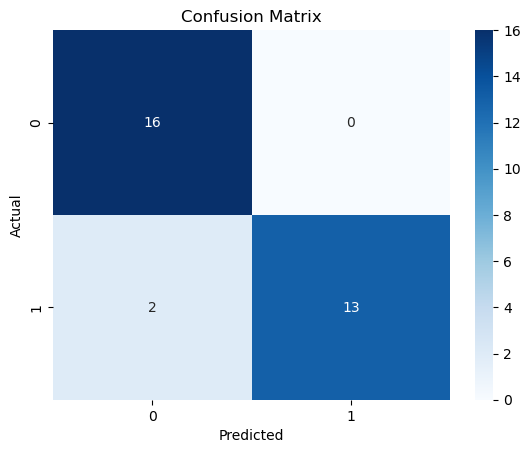}}
    \hfill
    \subfigure(B){\includegraphics[width=0.45\textwidth]{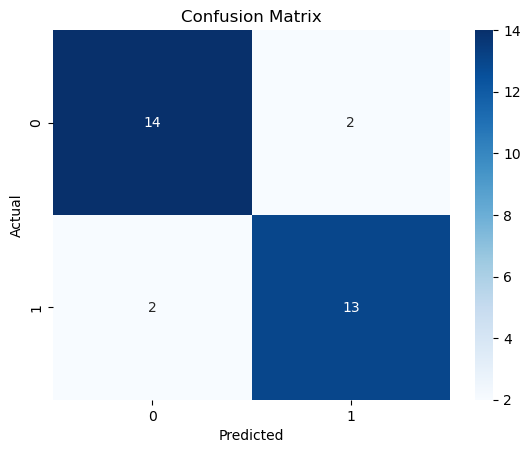}}
    \caption{Confusion Matrices showing predicted versus actual labels for Random Forest and Extra Trees. (A) Random Forest; (B) Extra Trees}
    \label{fig:my_label}
\end{figure}

\subsection{Statistical Analysis}
\indent Statistical significance was calculated for the features using Statsmodels (https://www.statsmodel.org/). The R² coefficient of determination— a measure representing the proportion of the variation in the dependent variable that can be explained by the independent variable— was calculated to be 0.87. The log-likelihood, a measure evaluating the goodness of fit of a model to a set of observed data— was computed to be 999.24. P-value statistics were calculated on a per feature basis. Out of a total of 1280 calculated features, 1010 features had p<0.001, 20 features had p<0.01, 29 features had p<0.05, and 181 features had p>0.05. \newline
\indent A correlation matrix was created to find how correlated each frequency band was to the diagnosis, while also providing pairwise relation between several frequency bands. Similar to a correlation matrix, feature importance was also calculated to find the importance of several factors in the Random Forest classifier prediction, the best performing model. Feature importance was calculated in two ways, the importance of each extracted feature and each filtered frequency band in the prediction. The results of the correlation matrix and feature importance are shown in Figure 7 and Figure 8.\newline

\begin{figure}[H]
    \centering
    \subfigure(A){\includegraphics[width=0.45\textwidth]{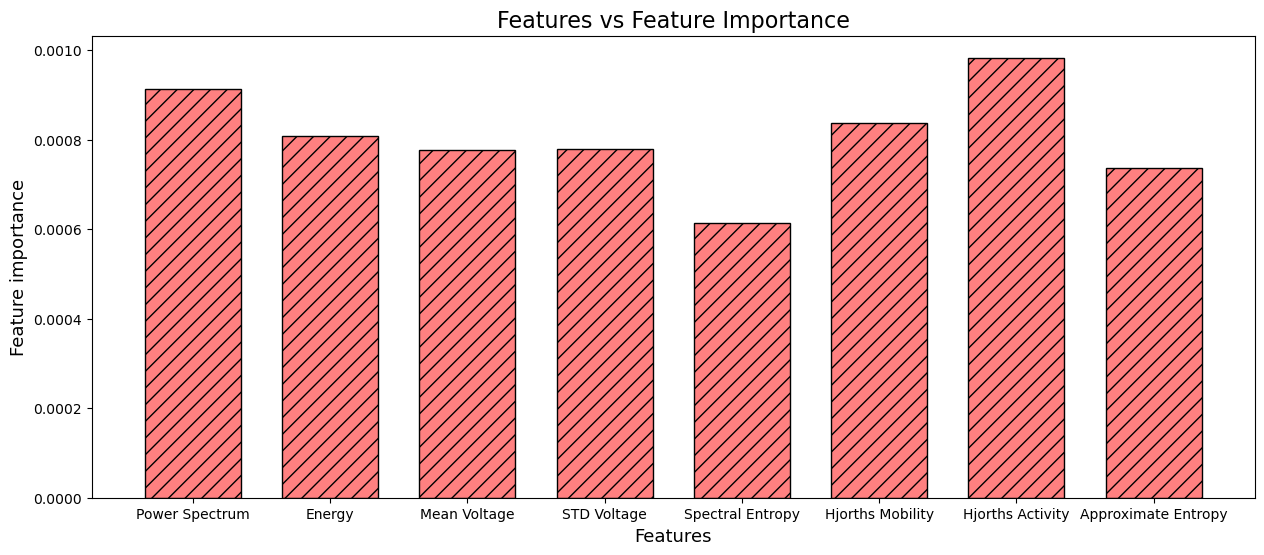}}
    \hfill
    \subfigure(B){\includegraphics[width=0.45\textwidth]{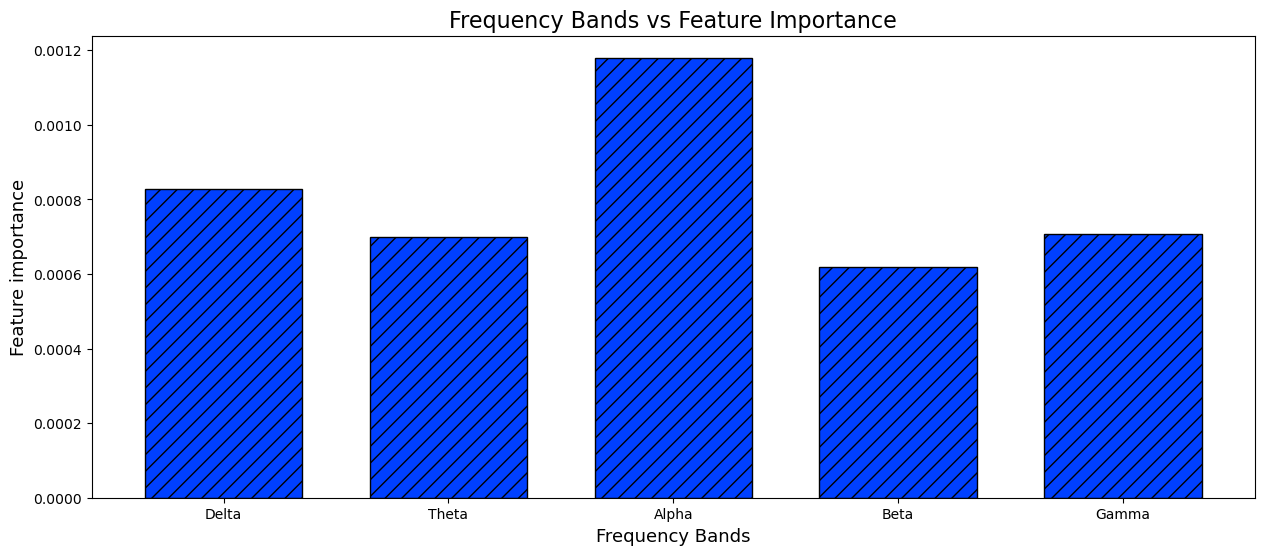}}
    \caption{Feature importance for each extracted feature and frequency band. (A) Features versus Feature Importance; (B) Frequency Bands versus Feature Importance}
    \label{fig:my_label}
\end{figure}

\begin{figure}[H]
    \centering
    \subfigure{\includegraphics[width=0.45\textwidth]{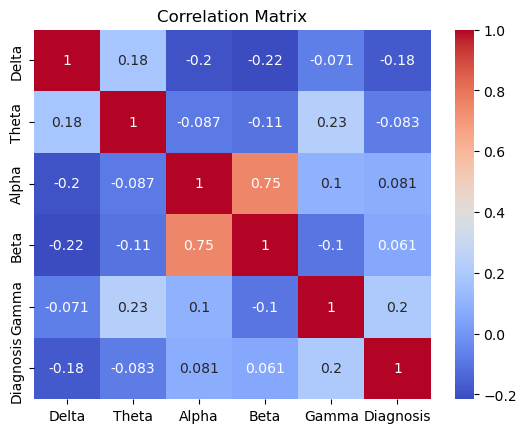}}
    \caption{Correlation matrix for each frequency band and the diagnosis}
    \label{fig:my_label}
\end{figure}

\section{Discussion}

\indent In this proof-of-concept study, we demonstrate that resting-state EEGs in the time, frequency, time-frequency, and nonlinear domains reveal significant differences between HC and PD individuals. Additionally, an explainable machine learning pipeline, guided by signal processing and statistical significance, results in classifier models that can accurately distinguish subjects as either HC or PD.\newline
\indent Parkinson’s is one of the most common causes of dementia, and despite some progress towards sustainable treatment, most interventions lack the ability to slow down disease progression due to improper diagnosis [38]. Assessment of brain neural activity could lead to a quantitative and more comprehensive diagnosis, thus delaying progression of Parkinson’s before major cognitive deficiency emerges.\newline
\indent Several EEG studies have been conducted to detect abnormalities in the brain function of Parkinson’s patients [8][39][40]. PD is characterized by neural activity in the 1 and 30 Hz frequency range in terms of tremor, one of the most popular effects of PD. Therefore, applying a band-pass filter across the 0.5 to 45 Hz range was appropriate. Characterizing PD requires a comprehensive representation of an EEG signal, which is why features were extracted in the time, frequency, time-frequency, and nonlinear domains after signal preprocessing and decomposition. Additionally, the choice of k=10 for 10-fold cross validation provided a more reliable estimate of the model’s performance along with reduced overfitting when partitioning the extracted features into training and test sets. \newline
\indent In this study, the features for HC or PD-afflicted individuals were statistically significant. 1059 out of the total 1240 features had p<0.05, suggesting a statistically significant model. 1010 features, the majority of all the features, had p-values <0.001, indicating a high level of confidence that most of the extracted features were not calculated by chance. For these features, we can safely reject the null hypothesis and determine that there is no significant evidence in support of the alternative hypothesis. \newline
\indent Since the main objective of this study was to find which features are most indicative of PD, feature importance was calculated for each individual feature and frequency band. In terms of feature type, Hjorth’s Activity (HA) yielded the highest feature importance, suggesting that future experimentation with feature extraction in the nonlinear domain can be explored. The observation that the alpha ($\alpha$) frequency band in the 8.5 to 11.5 Hz region had the highest feature importance is reasonable since previous research suggests that PD abnormalities are exposed in the 1 to 30 Hz region [20]. However, the lowest feature importance was calculated in the beta band, a frequency range of 15.5-30 Hz, which indicates some amount of noise in the beta band since it belongs to the frequency region where PD neural activity is apparent. Gamma’s frequency band, the only frequency region outside of the frequency for PD neural activity, had the second lowest feature importance, showing correlation to the prior observation on PD abnormalities in the 1 to 30 Hz range. Overall, the feature importance reflects a large amount of information about which feature algorithms are most suitable for differentiating between healthy individuals and PD patients. \newline
\indent The R² coefficient of determination showed that 87{\%} of variation in the dependent variable (PD diagnosis) can be explained by the independent variables (extracted features), which closely corresponds to the accuracy of the two best performing models, Extra Trees and Random Forest. \newline 
\indent The superiority of Random Forest and Extra Trees over the other two classifiers suggest that the ensemble tree method is suitable for binary classification after feature extraction in HC and PD individuals. The consistent use of Random Forest and Extra Trees analysis in previous literature suggests that the ensemble learning method, a collection of decision trees, outperforms a simple decision tree when classifying between HC and PD [41]. The results of this study provide significant performance improvement over Wang et al. (2021) who used Energy of EEG data and achieved 89.34{\%} classification accuracy for PD, suggesting that feature extraction in other domains besides time-frequency might be a better approach[42]. \newline
\indent There are several limitations in this study. First, Electroencephalography (EEG) is inherently noisy because of slight muscle movements and eye twitches even during a resting-state, and our analysis didn’t remove artifacts due to noise. Although our study still achieved significant metrics, artifact removal could provide more consistent results. Second, our study averaged channel data to improve the signal-to-noise ratio, but this approach could have simultaneously reduced dimensionality of the EEG data, resulting in a loss of possibly important spatial information. Third, this study didn’t assess EEG recordings at different stages of PD patients, which would have revealed several timeframe abnormalities useful when comparing Parkinson’s to HC.

\section{Conclusion}
\indent In this study we preprocessed, decomposed, and extracted eight features from four different domains for each Electroencephalographic signal. Then, four ML classifiers were used to accurately differentiate between healthy individuals and Parkinson’s patients. We found high ROC AUC with Extra Trees and Random Forest, two tree ensemble algorithms. After evaluating statistical significance on the extracted features and the results of the classifiers, we found the two most significant features to be Hjorth’s Activity (HA) and Power Spectral Density (PSD) and the most important frequency band to be the alpha ($\alpha$) band. While there is still much research to be done in this field, this study provides confidence in the use of Electroencephalography (EEG) for PD prognosis, and future research exploring more robust classifier models, feature extraction algorithms, preprocessing methods, and data cleaning techniques could build upon the experimentation results within this study. Topics for future research include exploring the comparison between HC and PD at different stages of Parkinson’s and the possible use of Magnetoencephalography (MEG), another technique for measuring brain electrical activity, to explore abnormalities in PD.

\end{document}